\documentclass[runningheads]{llncs}

\usepackage{eccv}

\usepackage{eccvabbrv}

\usepackage{graphicx}
\usepackage{amsmath}
\usepackage{amssymb}
\usepackage{enumitem}
\usepackage{algorithm}
\usepackage{algpseudocode}
\usepackage{tabularx}
\usepackage{makecell}
\usepackage{booktabs}
\usepackage{multirow}
\usepackage{tikz}
\usepackage{xcolor}
\usepackage{flushend}
\usepackage[accsupp]{axessibility}
\usepackage{bm}
\newcommand{\indc}[1]{{\mathbf{1}_{\left\{{#1}\right\}}}}

\newif\ifcomments
\commentsfalse

\newcommand{\nbo}[1]{\ifcomments{\sf\color{orange}[#1]}\fi}
\newcommand{\ls}[1]{\ifcomments\nbo{LS: #1}\fi}
\newcommand{\rd}[1]{\ifcomments\textcolor{blue}{[\small ry: ~#1~]}\fi}
\newcommand{\fei}[1]{\ifcomments\textcolor{red}{[\small fei: ~#1~]}\fi}
\newcommand{\ding}[1]{\ifcomments\textcolor{brown}{[\small ding: ~#1~]}\fi}

\usepackage[accsupp]{axessibility}  

\usepackage{hyperref}

\usepackage{orcidlink}
\usepackage{diagbox}

\definecolor{darkred}{RGB}{192, 0, 0}
\definecolor{darkblue}{RGB}{66, 131, 166}

\begin{document}

\title{Non-transferable Pruning} 


\author{Ruyi Ding\inst{1}\orcidlink{0000-0002-0079-8265} \and
Lili Su\inst{1}\orcidlink{0000-0003-3538-5679} \and
Aidong Adam Ding\inst{1}\orcidlink{0000-0003-1397-2442} \and
Yunsi Fei \inst{1}\orcidlink{0000-0002-9930-0868}}

\authorrunning{R. Ding et al.}

\institute{\email{\{ding.ruy, l.su, a.ding, y.fei\}@northeastern.edu}\\Northeastern University, Boston, MA 02115, USA
}

\maketitle
\begin{abstract}

\fei{what is the usage of the first line?  Generability of DNN is improved, and yet you would like to reduce it?}
\rd{Revise it, try to express that the model generability improves so that it will be vulnerable to be fine-tuned.}
\ls{Move your revised first sentence to line 14?}

Pretrained Deep Neural Networks (DNNs), developed from extensive datasets to integrate multifaceted knowledge, are increasingly recognized as valuable intellectual property (IP). 
To safeguard these models against IP infringement, strategies for ownership verification and usage authorization have emerged.
Unlike most existing IP protection strategies that concentrate on restricting direct access to the model, our study addresses an extended DNN IP issue: applicability authorization, aiming to prevent the misuse of learned knowledge, particularly in unauthorized transfer learning scenarios.
We propose \textbf{Non-Transferable Pruning} (NTP), a novel IP protection method that leverages model pruning to control a pretrained DNN's transferability to unauthorized data domains.
Selective pruning can deliberately diminish a model's suitability on unauthorized domains, even with full fine-tuning. 
Specifically, our framework employs the alternating direction method of multipliers (ADMM) for optimizing both the model sparsity and an innovative non-transferable learning loss, augmented with fisher space discriminative regularization, to constrain the model’s generalizability to the target dataset. 
We also propose a novel effective metric to measure the model non-transferability: Area Under the Sample-wise Learning Curve (SLC-AUC). This metric facilitates consideration of full fine-tuning across various sample sizes. 
Experimental results demonstrate that NTP significantly surpasses the state-of-the-art non-transferable learning methods, with an average SLC-AUC at -$0.54$ across diverse pairs of source and target domains, indicating that models trained with NTP do not suit for transfer learning to unauthorized target domains. 
The efficacy of NTP is validated in both supervised and self-supervised learning contexts, confirming its applicability in real-world scenarios.  \href{https://github.com/RollinDing/DNN-NTL-Pruning}{Git Repo}

\end{abstract}
\section{Introduction} \label{sec: intro}

Rapid advancements in deep learning, characterized by an increase in the model size, complexity, and capability, have made the training of deep learning models more time-consuming and data-intensive, 
yielding the pre-trained model a valuable intellectual property (IP)\cite{he2016deep}.
Protecting machine learning IP is therefore vitally important and yet challenging. 
Most existing IP protection strategies are reactive in nature, focusing on mitigating the damage afterwards rather than preventing breaches outright. 
These strategies 
detect unauthorized use of models by employing techniques such as watermarking~\cite{li2019prove} and fingerprinting~\cite{cao2021ipguard}, 
accompanied by legal procedures to penalize IP infringement. 

\begin{figure}[t]
    \centering
    \includegraphics[width=0.79\linewidth]{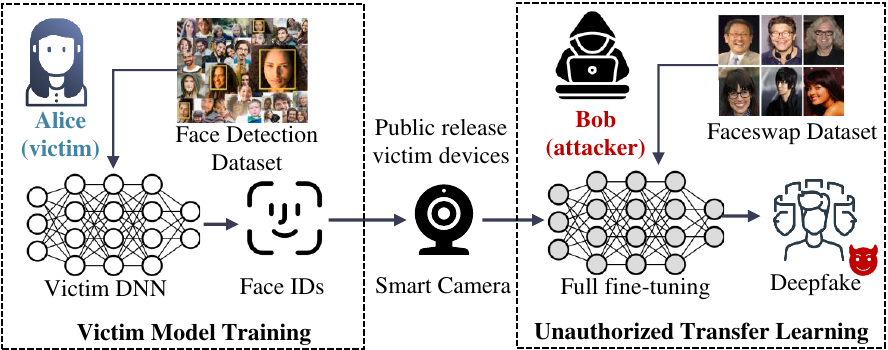}
    \caption{\textbf{Adversarial Scenario:} model vendor (Alice) built a DNN for face recognition (i.e., Face IDs) and distributed the smart camera (edge device) with this model onboard. 
    The malicious user (Bob) bought the device and obtained full access to the victim model. 
    He modified the model with transfer learning by fine-tuning with a target dataset to redirect the model for the malicious task (e.g., deepfake~\cite{westerlund2019emergence}). 
    } 
    \label{fig:case-study}
\end{figure}

Recently, in response to the limitations of passive protection methods, a new type of IP protection approach, known as Non-transferable Learning (NTL), is proposed, serving as a proactive measure to safeguard more general model IP~\cite{wang2021non, wang2023model}. 
Transfer learning has been employed to leverage the knowledge learned in pre-trained models towards new applications with effective fine-tuning. 
\fei{why did the previous line (commented out now) only mention computer vision? does transfer learning only help computer vision? also terminology-wise, what is the difference between (downstream) task, application, domain, and dataset?}
\rd{We focus on the computer vision task in this paper, may do not mention the computer vision here. Task=application, domain=dataset}
The learned knowledge, manifested in the pre-trained model structure and parameters, facilitates efficient fine-tuning, necessitating less amount of training data for the new task and bolstering the model performance. 
However, application of transfer learning also presents a threat to the model vendors--unauthorized tasks may be hostile, against the law, or simply cause economic disadvantages.
In response, NTL aims to restrict the misuse of a pretrained model, by controlling the inherent model transferability, thereby directly limiting its performance on unauthorized tasks. 
An example of this threat is depicted in \cref{fig:case-study}, showcasing how a model designed for benign purposes like face detection can be repurposed through transfer learning for a malicious task - Deepfake~\cite{westerlund2019emergence}.

\fei{the previous line is still verbose. do you need both "restrict the misuse" and "prevent the model exploitation."}
\rd{revised}
\ls{We shall discuss the needs of using the terminology "pretrained models" and "downstream applications."} 
\rd{Only keep the pretrianed model here}

Current NTL approaches~\cite{wang2021non, wang2023model} predominantly address scenarios where malicious users (attackers) have limited access to model parameters, preventing them from fully fine-tuning the victim model.
This paper, however, explores a more general and stronger attack scenario: the victim model is fully exposed, such as those deployed on edge devices as shown in~\cref{fig:case-study}. 
Such distinction is critical, as it means that attackers can directly access the pre-trained model and further manipulate it. 
With the growing trend of embedding machine learning services locally on edge devices, models are either directly available to attackers or can be easily extracted~\cite{zhu2021hermes, wei2020leaky, hu2020deepsniffer}. 
In these instances, attackers, equipped with unlimited access to model parameters, can potentially redirect the entire model towards unauthorized datasets.
Moreover, Non-transferable Learning~\cite{wang2021non} and Model Barrier~\cite{wang2023model} primarily use a specific objective function and gradient-based optimization.
Yet, we observe that these methods lack robustness to fully fine-tuning, as models may maintain target domain performance with sufficient training data and carefully selected hyperparameters.
Going beyond mere parameter optimization, preventing malicious transfer necessitates more substantial modifications to the DNN to further limit its capacity for the target domain.
Therefore, we leverage model pruning, which involves strategically zeroing out specific parameters to effectively constrain the model's transferability~\cite{fu2023robust}.
\fei{rephrase the previous line. what is exactly the limitation of previous NTL methods?} \rd{revised}
\fei{the introduction of pruning is very abrupt here. what are other alternative/possible methods to limit the model transferability?  why pruning? the motivation for using pruning for nontransferrable is missing}
\rd{Add and revise this part. But I feel like the motivation is caused by the stronger attacker fully control the model", which may a little bit duplicate to the previous paragraph.}

We propose \textbf{Non-transferable Pruning (NTP)}, a novel model IP protection mechanism designed to nullify adversaries' transfer learning attempts. 
Our approach leverages a specialized pruning technique, employing the Alternating Direction Method of Multipliers (ADMM), aimed at selectively diminishing the model's transferability to the targeted domain while preserving its efficacy within the source domain. 
We also propose a novel metric, the Area Under the Sample-wise Learning Curve (SLC-AUC), to assess model transferability. 
This metric extends beyond the previous model non-transferability measurement in~\cite{wang2021non, wang2023model} using only the accuracy degradation in the target domain.
SLC-AUC evaluates the volume of fine-tuning data required by an attacker to redirect the model for the target domain, and compares the fine-tuning performance of the compromised model against the one trained from scratch.
The new metric is more sound and provides a thorough evaluation of how pre-learned knowledge from the victim model helps fine-tune effective models for the target domain.

In summary, our contributions of this work are three-fold:
\begin{itemize}
    \item We propose the first ADMM-based model pruning strategy that focuses on limiting the model non-transferability, so as to protect the intellectual property of the pretrained DNN model.
    We also propose a Fisher Space Regularizer which penalizes class discrepancy on the target domain during NTP to enhance the method's robustness under a fully fine-tuning attack scenario.
    \item We propose a new evaluation metric, Area Under Sample-wise Learning Curve (SLC-AUC), to measure model non-transferability, considering the variation of model performance with different training data volume
    from the target domain.
    It overcomes the bias of prior metrics, such as accuracy degradation, which do not address the different model transferability when the attacker has a different amount of target samples. 
    \item We evaluate our NTP on a vast range of datasets and models. Compared to the SOTA NTL  methods~\cite{wang2021non, wang2023model}.
    The NTP shows more transferability reduction (the SLC-AUC scores are -$0.54$ on average).  
    We also evaluate NTP on large-scale datasets (subsets of ImageNet) and self-supervised models (SimCLR~\cite{chen2020simple}, MoCo~\cite{he2019moco, chen2020improved}), demonstrating the wide applicability of NTP.
\end{itemize}

\section{Related Works} \label{sec: background}

\subsection{Model IP Protection and Non-transferable Learning} \label{bg: ntl}
Model IP protection aims to protect well-trained DNN models, with typically two types of techniques: ownership verification and usage authorization.
Ownership verification is a passive defense against IP infringement, showing evidence of the illegal model usage by an attacker.
Commonly used ownership verification methods include DNN watermarking~\cite{li2019prove, song2017machine, wu2020watermarking, guo2018watermarking} and model fingerprinting~\cite{cao2021ipguard, chen2019deepmarks, zhao2020afa}.
However, watermark removal attack~\cite{guo2020fine, hitaj2018have, pan2023cracking} can evade such protections. 
In contrast, usage authorization aims to restrict the user access of the model, such as the prior works that lock the neural network with encryption~\cite{alam2020deep, chakraborty2020hardware} where only the authorized user with specific keys can unlock the model. 
Yet, this method has its drawbacks: authorized users with the keys can fully access and potentially misuse the model, presenting another form of IP infringement.
\ls{any limitation of such encryption?} \rd{I want to mention something related to side-channel attack here so that the encrypted model may still be vulnerable to model stealing attack, However, currently I follow a claim more related to NTL}

Non-transferable Learning extends the definition of model usage--not only the direct use of the model should be authorized, but also further usage of the model should receive some restriction, known as applicability authorization~\cite{wang2021non, wang2023model}.
Specifically, given a target domain on which the model is not allowed to apply, NTL aims to restrict the model transferability towards it, denoted as $\mathcal{L}_{NTL} = \mathcal{L}_\mathcal{S}+ \mathcal{R}_\mathcal{T}$,
where $\mathcal{L}_\mathcal{S}$ ensures model efficiency in the source domain and $\mathcal{R}_\mathcal{T}$ signifies the NTL penalty on unauthorized target domains. 
Wang et al.~\cite{wang2021non} use a combination of Kullback-Leibler divergence and Maximum Mean Discrepancy for this penalty;
Compact Un-Transferable Isolation (CUTI)~\cite{wang2023model} enhances the non-transferability by differentiating between shared and private features of the source and target domains.
However, existing studies primarily assess non-transferability with accuracy degradation through direct inference on the target domain, neglecting scenarios where the model undergoes fine-tuning.
Our research addresses this gap by examining an attack scenario wherein a malicious user can fully fine-tune the victim model. 
We introduce a novel metric for evaluating transferability across various data scales and employ DNN model pruning in NTL, yielding superior results compared to existing approaches.
\ls{English needs to be polished up a bit. Pay attention to the consistency of terminology. Avoid using multiple terms to refer to the same thing. }
\rd{revise the language here.}

\subsection{Weight Pruning} \label{bg: admm}
Weight pruning~\cite{zhu2017prune, he2017channel, he2019filter}, as a deep learning model compression technique, has been used for deploying DNN on resource-constrained platforms, exploiting weight sparsity to trim down neuron connections without notable accuracy degradation.
Mainstream pruning methods can be broadly categorized into two types: structural pruning~\cite{ding2019centripetal, filterspruning, liu2021group, you2019gate} and unstructural pruning~\cite{dong2017learning, lee2019signal, sanh2020movement}.
Structural pruning focuses on removing neurons or channels, while  unstructured pruning zeros out specifically chosen weights.
To adapt model pruning for DNN non-transferable learning, we find unstructural pruning is suitable and succeed in preserving the model performance in the source domain.
Model pruning tasks can be formalized as a dual-optimization problem considering both performance and model sparsity, and solved with the alternating direction method of multipliers (ADMM), which has been used in model adversarial pruning~\cite{ye2019adversarial, gui2019model, jian2022pruning, ye2018progressive}, where the pruning scheme aims to achieve the DNN adversarial robustness and model sparsity at the same time.
Model pruning is also used in DNN watermarking~\cite{xie2021deepmark, zhao2021structural}, and the pattern of pruning can embody owner's fingerprints~\cite{zhao2020afa}.\fei{not sure if I understood your point in the previous line and rephrased correctly}
In this work, we leverage ADMM-based pruning methods for applicability authorization, which achieves model non-transferability via weight pruning and updating, resulting in target domain transferability reduction of a compact model, which shows strong robustness against model fine-tuning.
\fei{the last line is perpexing.  there is no "not only" and "but also."  Everything is regarding nontransferrability - only one thing.  also, what is exactly "capacity?" change a more clear term or at least define it early on}
\rd{revised}
\fei{rephrase the previous line} \rd{revised. Just to introduce ADMM here.}
\fei{rephrase the previous line.??? what IP protection strategies? fine-tuning and pruning are watermark removal attacks}
\rd{remove this line, do not think we need to mention it here.}

\section{Threat Model} \label{sec: threat-model}
\noindent\textbf{Adversarial Goals}: 
We have a victim model designed for a specific task, on a \textit{source domain}. 
The attacker aims to use this victim model to create a new model for a malicious task, on a \textit{target domain}. 
The attack hinges on leveraging the transferable knowledge from the source domain within the victim model to efficiently develop a model for the target domain.

\noindent\textbf{Attacker Profile}:
We assume the attacker possesses complete control of the pre-trained DNN, knowing the model's structure and parameters in full detail.
The attacker also has access to a subset of the unauthorized target domain dataset, and can fully fine-tune the model for the target domain with sufficient computational resources. 
Potential adversaries could be end users of DNN models deployed on edge devices, malicious administrators within cloud-based environments, or external attackers capable of model extraction.

\noindent\textbf{Defender Capability}:
Our defense strategy is from the perspective of the victim model provider, which proactively safeguards the model in insecure environments. 
The primary goal is two-fold: in the source domain, the model is legitimately used for the original task with good performance; in the target domain, the pretrained model does not offer any transferable knowledge to yield a fine-tuned model that outperforms the model developed from scratch (without pretrained model).  
Given that transfer learning typically requires adjustments to the linear classifiers due to discrepancies in label space sizes, NTP concentrates on modifying the feature extractor in the pre-trained models, which is integral to determining how the model interprets and processes input data. 

\fei{why do you need to mention this comparison?}
\rd{Indeed I am not sure about it. We may need to mention it only in SP paper not here.}

\ls{Justify why not consider this? For example, they require speical hardwares?} \rd{I mention this here because the reviewer from SP talk about such attacks, which is out of the scope of our research. 
Not sure if we need to mention this in this work or not.}

\ding{We need to clarify what transfer-learning tasks on target domain is aimed at. Are we using the same model on source domain and just adjust its weights for new target domain task? Or the model on source domain as an encoder (feature extraction) and add task-specific classifier after? The later experiments seem to contain both scenarios. The framework here has to be general and precise to cover both if that is what you mean.}
\rd{Our framework basically is focus on the encoder part, as we consider the classifier part might be changed due to the different size of feature space. However, we also show that our work is robust to change the classifier. }
\section{Model Non-transferable  Pruning} \label{sec: method}
\begin{figure}[t]
    \centering
    \includegraphics[width=0.88\linewidth]{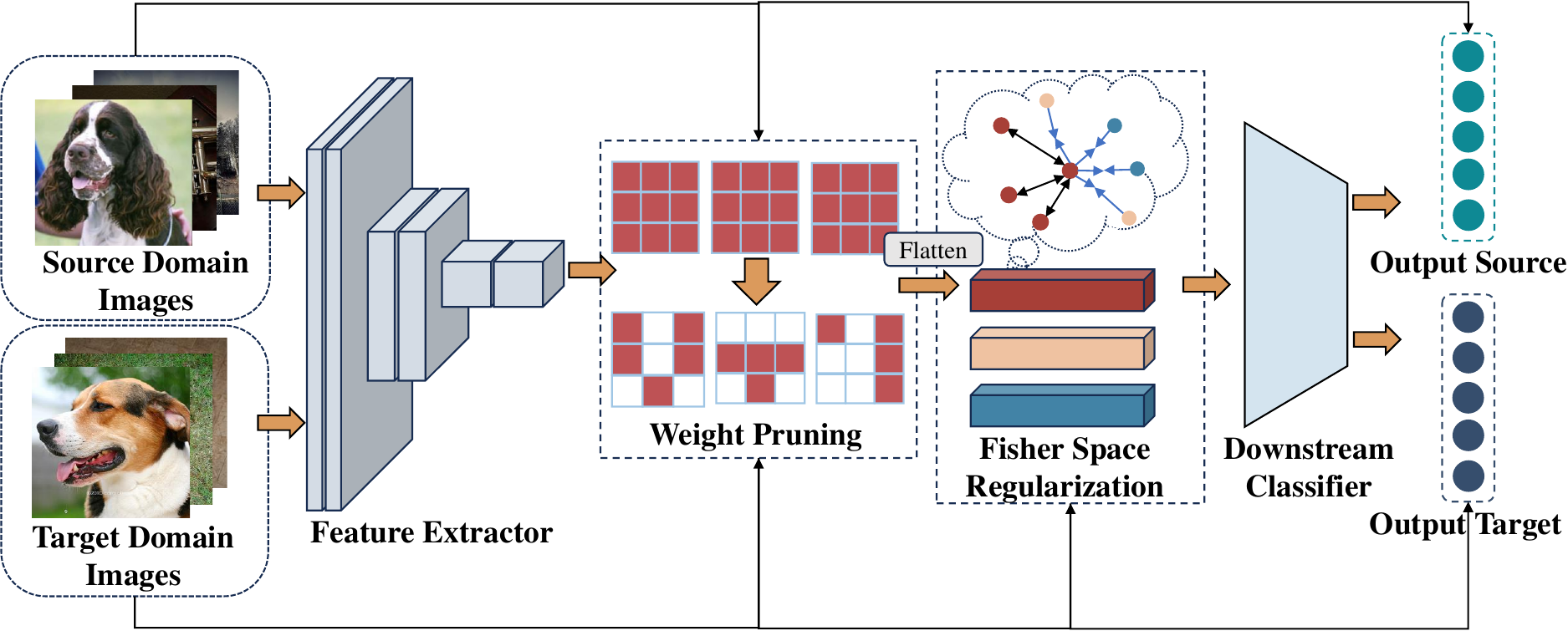}
    \caption{An Overview of Non-transferable Pruning Procedure}
    \label{fig: pipeline}
\end{figure}

In this section, we illustrate the proposed novel DNN applicability authorization method, Non-Transferable Pruning (NTP).  \cref{fig: pipeline} depicts an overview of the procedure. NTP is informed by data from both the source and the target domain, 
to selectively prune model parameters that are crucial only to the target domain, thereby diminishing the model's transferability to it.
We use the 
ADMM-based method for weight pruning. 
Furthermore, we implement a new regularization technique, based on Fisher discriminant analysis in the feature space, to further reduce the transferability even after fine-tuning. 
This technique adjusts the target domain feature space by reducing inter-class distances and increasing intra-class distances, which effectively blurs decision boundaries. 

\subsection{Model Non-transferability Metric} \label{method: measurements}

\begin{figure}[t]
    \centering
    \includegraphics[width=0.99\linewidth]{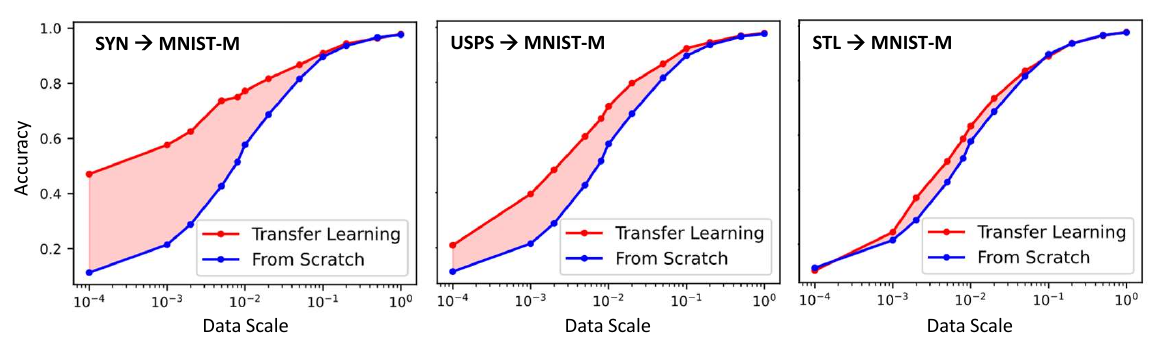}
    \caption{\textbf{Sample-wise Learning Curve and Area Under Curve}: The target domain is MNIST-M~\cite{ganin2016domain} and the pretrained model is trained with network architecture ResNet18~\cite{he2016deep} on datasets SYN~\cite{ganin2015unsupervised}, USPS~\cite{hull1994database}, and STL~\cite{coates2011analysis}, respectively. }
    \label{fig: method: SLC curve}

\end{figure}

Previous non-transferable learning research primarily assesses non-transferability by measuring the accuracy degradation in target domains~\cite{wang2021non, wang2023model}. 
Their evaluations are typically confined to fixed transfer learning settings, especially the sample size for fine-tuning, which may not adequately capture the complexity and variability inherent in real-world malicious transfer learning scenarios, especially those where attackers have unrestricted access to both the model and target dataset~\cite{zhuang2020comprehensive}.
To address this gap, our work broadens the evaluation scope by considering 
various training data proportions, offering a more comprehensive and dynamic assessment of model non-transferability.
We propose the Area Under the Sample-wise Learning Curve  (SLC-AUC) as a new metric for nontransferrability.  
This metric evaluates the performance of a pretrained model transferred to the target domain across diverse scales of samples~\cite{sun2017revisiting}. On an SLC plot, the X-axis (logarithmic scale) is the size of the target sub-dataset for fine-tuning or learning, and the Y-axis is the testing accuracy of the model.  There are two curves in the plot, one for the fine-tuned model (transferred one, in {\color{red}red}) and the other for the model without any prior knowledge (learned from the sub-dataset from scratch, i.e., with random initial parameters, in {\color{blue}blue}). 
\begin{definition}[SLC-AUC metric]
For any given pretrained model, its SLC-AUC is defined as the difference (the \textit{shaded} area) between the two SLCs (one for the transferred model and the other for the model trained from scratch).  
\end{definition}
A positive SLC-AUC value indicates effectiveness of transfer learning; while a negative SLC-AUC value means non-transferrability.  
\cref{fig: method: SLC curve} shows that remarkably, initiating training with a pretrained model can yield superior performance for different pair of source and target domains with a ResNet18 model, particularly when dealing with similar datasets.
This aligns with the intuition behind transfer learning, where users can leverage knowledge encoded in a trained model to guide the model fine-tuning for a new task despite limited samples. 
However, when an attacker has access to abundant data, he can often obtain a model, based solely on the target data without relying on external knowledge, as effective as the one transferred from the pre-trained model. 
Intuitively, the higher SLC-AUC value, the larger transferability from the source domain to the target domain. 
~\cref{fig: method: SLC curve} evaluates the model transferability for three different pairs. 
From SYN (synthetic digits dataset) to MNIST-M (digits with background) has the largest SLC-AUC, indicating that a model pre-trained on SYN can generalize well to MNIST-M with only a few target samples.
On the contrary, the model trained on STL (image recognition dataset) provides no apparent advantage in fine-tuning for the MNIST-M dataset, bearing the smallest SLC-AUC.

Our goal is to design a model pruning procedure with non-transferable objective so that the pre-trained model has a small or even negative SLC-AUC for selected target domains. In other words, the performance of fine-tuned models on the target dataset are comparable to, or lower than, a direct-trained model.

\subsection{Exploring DNN Model Sparsity for Transferability} \label{method: observation}

DNNs are often over-parametric. Model pruning aims to identify a compact subnetwork 
by eliminating redundant parameters for efficiency and performance. 
Recent studies find that the compact structure of a pruned model 
might significantly impair its transferability~\cite{liu2018rethinking, zhu2017prune, chen2020lottery, frankle2018lottery, chen2021lottery}. 
Intuitively, this may be because a compact model only contains the information that is most relevant to the source domain. 
There are recent efforts to enhance the transferability of pruned models~\cite{fu2023robust, liu2021transtailor} using adversarial examples or task-specific pruning. 
In this paper, we try to answer 
an intriguing opposite question: \textit{can model pruning be leveraged to identify a compact DNN model that is effective in the source domain but less effective, or even adversarial, when transferred to the target domain?}\fei{is it DNN "model" or DNN "structure". also, "in terms of transferability" seems to be inappropriate here} \rd{revised the last sentence. DNN "model" might be better as the pruning not only find the structure but also has some optimization on the parameters.}

\begin{figure}[t]
    \begin{minipage}[t]{0.49\textwidth}
    \centering
    \includegraphics[width=\linewidth]{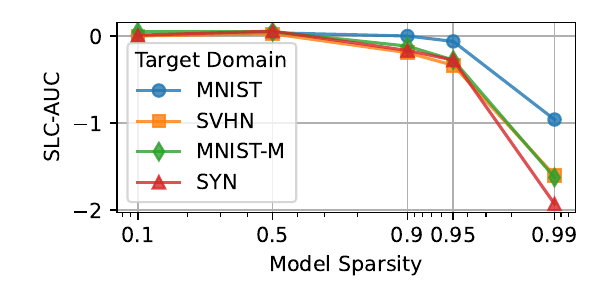}
    \caption{Sparsity VS SLC-AUC}
    \label{fig: sparsity-motivation}
        \end{minipage}
    \hfill
    \begin{minipage}[t]{0.47\textwidth}
        \centering
    \includegraphics[width=\linewidth]{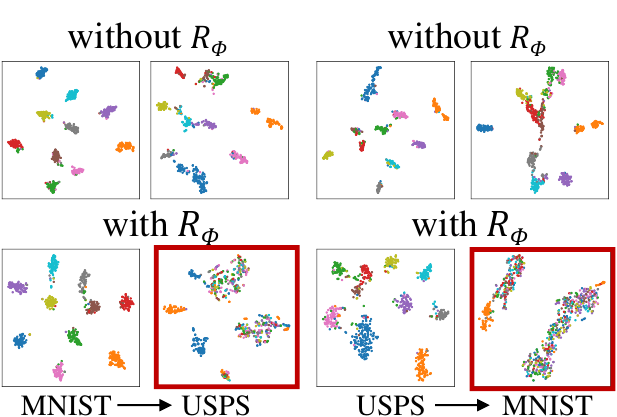}
    \caption{Feature spaces visualization}
    \label{fig: feature-space}    
    \end{minipage}
\end{figure}

We analyze the impact of model sparsity on its transferability through a toy example, and the results are shown in \cref{fig: sparsity-motivation}. 
We pretrain a ResNet-18 model on the USPS dataset (the source domain), apply one-shot magnitude pruning~\cite{chen2021lottery} to it with different levels of sparsity, and measure the pruned models' transferability to different target domains. 
Our findings reveal the relationship between model sparsity and transferability. 
Specifically, models with high levels of sparsity exhibit lower SLC-AUC scores, indicating a diminished capacity for transferring due to a constrained parameter set. 
Conversely, models with mild pruning ratios sometimes experience enhanced transferability due to reduced overfitting on the target domain~\cite{liu2017sparse}. 
This observation from \cref{fig: sparsity-motivation} 
confirms our conjecture that a heavily-pruned network's capacity for transferring to unauthorized target domains can indeed be reduced. 
The naive one-shot pruning on weights, although straightforward, leads to 
severe accuracy degradation on the source domain.
In the following sections, we delve into designing an advanced pruning scheme to balance the non-transferability and the performance on the source domain. 

\subsection{Formulating Non-transferable Pruning Objective} \label{method: objective}

We consider a pretrained neural network, $f_{\mathbf{W}}$, designed for the source domain $\mathcal{S} = {(x_{\mathcal{S}}, y_{\mathcal{S}})}$.
We aim to reduce the model transferability to a target domain  $\mathcal{T} = {(x_{\mathcal{T}}, y_{\mathcal{T}})}$.
Without loss of generality, the network can be viewed as two parts: a feature extractor $\Phi_{\mathbf{W_1}}$ and a task classifier $\Omega_\mathbf{W_2}$, where $W_1$ and $W_2$ denote the feature extractor and classifier's trainable parameters, respectively.
Following the design flow of Wang et al.~\cite{wang2021non}, we formulate the non-transferable learning objective function in~\cref{eq: ntp-objective}: 
\begin{equation}
    \mathcal{L}_{\textit{NTP}}(\bm{W}) \triangleq \mathcal{L}_{\mathcal{S}}(\Omega_{W_2}, \Phi_{W_1}, \mathcal{S}) + \mathcal{R}_{\mathcal{T}}(\Omega_{W_2}, \Phi_{W_1}, \mathcal{T}) + \mathcal{R}_{\Phi_\mathcal{T}}(\Phi_{W_1}, \mathcal{T}), 
    \label{eq: ntp-objective}
\end{equation}
where $\mathcal{L}_\mathcal{S}$ is the cross-entropy loss of the source domain, $\mathcal{R}_\mathcal{T}$ is a regularization term that penalizes the performance on the task domain, and $\mathcal{R}_{\Phi_\mathcal{T}}$ is our newly introduced fisher space regularization term used to collapse features on target domain. 
A common choice of $\mathcal{R}_\mathcal{T}$ is $\mathcal{R}_\mathcal{T}=-\text{min}(\beta, \alpha\mathcal{L}_\mathcal{T})$, where $\alpha$ and $\beta$ are hyperparameters introduced to ensure the stability of the loss~\cite{wang2021non, wang2023model}. 
We notice that only the first two terms in~\cref{eq: ntp-objective} are not sufficient for ensuring non-transferability, as large target losses can arise from either incorrect class mapping or indistinct features, with the former still offering useful information to an attacker through fine-tuning. 
Hence, to enhance non-transferability, we introduce $\mathcal{R}_{\Phi_\mathcal{T}}$ for additional regularization in the model's feature space.

\ding{Please revise to add some statements of insight: previous work has $\mathcal{R}_\mathcal{T}$ term which leads to worse accuracy on the target domain. For the pre-trained classifier: the lower accuracy can be due to either (a) the features (outputs of encoder and inputs of classifier) were mapped to wrong classes; (b) the features distributions among different classes are similar thus hard to distinguish. We add $\mathcal{R}_{\Phi_\mathcal{T}}$ term to further encourage (b) occurrence more. The fine-tuning of encoder requires the separation among classes. As (b) encourages smaller class separations than random initiation, the pre-trained encoder is thus a bad initiation than random weights. The following theoretical analysis confirms this.   }
\rd{I got your point. Revised.}

\subsubsection{Fisher Space  Regularization:} \label{method: penalty}
Fisher Discriminant Analysis (FDA), denoted by $\mathcal{R}_{\Phi_\mathcal{T}}$, captures
class separability; the more separable the classes the stronger transferability~\cite{shao2022not}. 
It is formally defined as  
\begin{equation}
\label{eq: fisher regularizer} 
    \mathcal{R}_{\Phi_\mathcal{T}} \triangleq \gamma\frac{\sum_{c\in \mathcal{C}} ||\overline{z}_c -\frac{1}{|\mathcal{C}|}\sum_{c' \in \mathcal{C}} \overline{z}_{c'}||_2}{\sum_{c\in \mathcal{C}}\sum_{(x_i, y_i)\in \mathcal{T}} ||z_i -\overline{z}_c||_2\indc{y_i=c}}, 
\end{equation}\fei{What is c' in your equation (3)? In the denominator you enumerate all the labels, the nominator doesn't seem right.  do you need to have a Summation for all c as well? }
\rd{I see your point. The nominator is wrong here.}
where $\mathcal{C}$ is the label set of the target domain, 
$z_i = \Phi_{W_1}(x_i)$ is the extracted feature of $x_i$, 
$\overline{z}_c = \frac{1}{N_c}\sum_{(x_i, y_i)\in \mathcal{T}} z_i\indc{y_i=c}$ is the averaged features of all the samples whose label are class $c$\footnote{Here, $\indc{y_i=c}$ is an indicator function whose value is 1 if $y_i=c$ and 0 otherwise.},
and $\gamma$ is a regularization coefficient.
Intuitively, $\mathcal{R}_{\Phi_\mathcal{T}}$ quantifies how well the extracted features of the target domain dataset are clustered. The larger the difference of means between classes and the smaller the feature variance within each class, the better the cluster structures.
Minimizing $\mathcal{R}_{\Phi_\mathcal{T}}$ optimizes the feature space in a contrasting way, where we try to collapse features of data from different classes. 

Next, we verify this conjecture both numerically and theoretically.
We compare the feature space of learning with/without the fisher space regularization in~\cref{fig: feature-space}.
We visualize the feature spaces with t-SNE~\cite{van2008visualizing} between MNIST and USPS datasets. 
It demonstrates that $\mathcal{R}_{\Phi_\mathcal{T}}$ blurs class clusters in the target domain (in \textcolor{red}{red} box) yet maintains their discrepancy in the source domain.  
For simplicity, we present our theorem assuming $|\mathcal{C}|=2$, i.e., a binary classification problem, with the extension to many more classes being done analogously. 
\ls{bring the theory statement here?}
\rd{Do you mean only the statement and proof in the appendix? I feel like we have no enough space to write the proof here}\ls{only statement}
\begin{theorem}
\label{thm: main}
Let $\mathcal{T}$ be a given target domain. 
Suppose that its label space $\mathcal{C}=\{0, 1\}$. 
Let $C_0 = \{(x_i, y_i): y_i=0\}$ and $C_1 = \{(x_i, y_i): y_i=1\}$.
Suppose that $|C_0| = |C_1|$.   
Then there exists a neural network with a feature extractor $\Phi_{W_1}$ that minimizes Eq.\,\eqref{eq: fisher regularizer} and the distributions of the extracted features of classes 0 and 1 
are indistinguishable. 
\end{theorem}
\rd{Move the last statement out of Theorem 1.}

The proof of Theorem \ref{thm: main} 
can be found in Appendix A.
Theorem \ref{thm: main} is stated for the special scenarios where the numbers of samples of class 0 and class 1 are equal. Its analysis can be extended to more general settings yet.\fei{what general settings? unbalanced class size or many more classes?} \rd{many more classes, we simplify with two classes.}
Experiments in~\cref{exp: supervised model} show the effectiveness of $\mathcal{R}_{\Phi_\mathcal{T}}$ in reducing the feature's class discrepancy to affect the fine-tuning process.


\if 
We illustrate the conceptual correctness of $\mathcal{R}_{\Phi_\mathcal{T}}$ functionality from the view of a binary logistic regression model.
Assume a binary logistic regression model $\hat{y}=\sigma(\mathbf{W}^T\mathbf{x}+b)$, where $\sigma$ is the sigmoid function.
The cross-entropy loss can be written as,
$$\mathcal{L}(y, \hat{y})=-[ylog(\hat{y})+(1-y)log(1-\hat{y})]$$
Without loss of generality, we assume the model parameters are fine-tuned through gradient-based optimizers, i.e., stochastic gradient descent.
The gradient of loss with respect to weights $\mathbf{W}$ for sample $i$ is,
$$\nabla_\mathbf{W} \mathcal{L}^{(i)} = (\hat{y}^{(i)} - y^{(i)})x^{(i)}$$
The penalization term $\mathcal{R}_\mathcal{T}$ penalizes the model classification outputs so that the current model always makes a wrong prediction.

\ding{We can consider the classification problem starting with the features $z_i$ as inputs. It is known that DNN locally is approximately linear. The Fisher Regularization is specifically designed to make it hard for linear discrimination, which means the gradient for the DNN classifier with input $z_i$'s are going to be small. It is hard to train a good classifier using $z_i$ as inputs. Now, the fine-tuning also involving fine-tuning the weights in the encoder before the features outputs $z_i$. Those gradient adjustment for encoder weights come from back-propagation. Thus when gradients from the later classifier are very small, also leads to small back-propagated gradients on encoder weights, making the fine-tuning very slow.}
\rd{I remove the proof here to the appendix}

Therefore, given two samples $i, j$ from different classes, \cref{eq: fisher regularizer} regularize that $x^{(i)}\approx x^{(j)}$, so that the gradient $\nabla_\mathbf{W}\mathcal{L}^{(i)} \approx -\nabla_\mathbf{W}\mathcal{L}^{(j)}$, which causes a small gradient.
On the other hand, when sample $i, j$ are from the same class, \cref{eq: fisher regularizer} leads to more variability in the sign and magnitude of $x^{(i)}, x^{(j)}$. 
The gradient $\nabla_\mathbf{W}\mathcal{L}^{(i)} \cdot \nabla_\mathbf{W}\mathcal{L}^{(j)} = (\hat{y}-y)^2 x^{(i)}x^{(j)} \ll (\hat{y}-y)^2 ||x^{(i)}||\cdot||x^{(j)}||$ as a larger variance will cause a high cancellation effect due to different directions and magnitude of samples from the same class. 
In conclusion, for a binary logistic regression model, \cref{eq: fisher regularizer} can generate a smaller gradient during model fine-tuning. 
This conclusion can be extended to fine-tune a deep neural network:
$$\mathbf{W}'=\mathbf{W}-\eta\nabla_\mathbf{W} \mathcal{L}_\mathcal{T}$$ when fine-tuning the model with a learning rate $\eta$, we restrict the model generalizability:  via penalizing $\mathcal{R_\mathcal{T}}$, we ensure that current model will also give a wrong prediction on the target domain;
with \cref{eq: fisher regularizer}, we further against the fine-tuning on the target domain as the features space produces high intra-class variance and low inter-class variance features.
\fi

\subsection{Integrating Model Pruning with NTP Objective} \label{method: concurrent}
\begin{algorithm}[t]
\caption{Non-transferable Pruning}\label{alg:admm_pruning}
\begin{algorithmic}[1]
\State \textbf{Inputs:} Initial DNN weights $\bm{W}^{(0)}$, sparsity regularization parameter $\lambda$, 
\State \hspace{\algorithmicindent} penalty coefficient $\rho$, desired sparsity $S$, source domain $\mathcal{S}$, target domain $\mathcal{T}$  

\vspace{0.5em}
\State \textbf{Initialization:} $\mathbf{Z}^{(0)} = \mathbf{W}^{(0)}$, $\mathbf{U}^{(0)} = \mathbf{0}$, t=0
\While{$\text{card}(\mathbf{W}^{(t)})/|\mathbf{W}^{(t)}| < 1-S$}
    \Statex  \hspace{\algorithmicindent} \textcolor{gray}{/*$\mathbf{W}$-update (update model weight)*/}
    \State $\mathbf{W}^{(t+1)} = \arg\min_{\mathbf{W}} \left( \mathcal{L}_{\textit{NTP}}(\mathbf{W}) + \frac{\rho}{2} \|\mathbf{W} - \mathbf{Z}^{(t)} + \mathbf{U}^{(t)}\|_2^2 \right)$
    \Statex \hspace{\algorithmicindent}  \textcolor{gray}{/*$\mathbf{Z}$-update (update auxiliary variable)*/}
    \State $\mathbf{Z}^{(t+1)} = \arg\min_{\mathbf{Z}} \left( \lambda \|\mathbf{Z}\|_1 + \frac{\rho}{2} \|\mathbf{W}^{(t+1)} - \mathbf{Z} + \mathbf{U}^{(t)}\|_2^2 \right)$
    \Statex \hspace{\algorithmicindent} \textcolor{gray}{/*$\mathbf{U}$-update (update dual variable)*/}
    \State $\mathbf{U}^{(t+1)} = \mathbf{U}^{(t)} + \mathbf{W}^{(t+1)} - \mathbf{Z}^{(t+1)}$
    \State $t = t +1$
\EndWhile
\State \textbf{Output:} $\mathbf{W}$ 
\end{algorithmic}
\rd{Wondering do we need to add the upper subscription to W and Z for previous round here?}
\end{algorithm}

We use the ADMM-based model pruning strategy~\cite{zhang2018systematic, ren2019admm}.  
To obtain a NTP pre-trained model, we solve the following constrained optimization problem:
\begin{equation}
    \underset{\textbf{W}}{\text{minimize}} \quad \mathcal{L}_{\textit{NTP}}(\mathbf{W}), \quad
    \textbf{subject to} \quad \text{card}(\mathbf{W})/|\mathbf{W}| < 1-S
    \label{eq: ntp-pruning-objective}
\end{equation}
where $\text{card}(\cdot)$ returns the number of nonzero elements and $S$ is the desired model sparsity.
In particular, we realize the optimization 
via ADMM method~\cite{zhang2018systematic, ren2019admm}. 
It reformulates the original problem by introducing an auxiliary variable, denoted as $\mathbf{Z}$, and a dual variable $\mathbf{U}$ as an augmented Lagrangian.
The algorithm solves two primal objectives separately.
The first subproblem optimizes the weights $\mathbf{W}$ by minimizing the original non-transferable pruning loss function augmented with a penalty term to encourage $\mathbf{W}$ to be close to $\mathbf{Z}$ with coefficient $\rho$;
this unconstrained problem can be solved with stochastic gradient descent.
The second subproblem applies sparsity to the auxiliary variable $\mathbf{Z}$ through $L_1$-norm  regularization, which can be solved efficiently by applying soft thresholding.
Finally, we update the dual variable $\mathbf{U}$ based on the discrepancy between $\mathbf{W}$ and $\mathbf{Z}$.
The overall NTP framework is summarized in~\cref{alg:admm_pruning}.

\section{Experiments} \label{sec: experiments}
\subsection{Experiment Setup} \label{exp: settings}
\noindent\textbf{Datasets:} \label{exp: dataset}
Following previous works~\cite{wang2021non, wang2023model}, we evaluate our method on popular domain adaption datasets. Specifically, MNIST (MT)~\cite{deng2012mnist}, USPS (US)~\cite{hull1994database}, SVHN (SN)~\cite{netzer2011reading}, MNIST-M (MM)~\cite{ganin2016domain} and SYN (SD)~\cite{ganin2015unsupervised}, are commonly used digits dataset containing digits from 0 to 9 from varies scenes;
CIFAR-10~\cite{krizhevsky2009learning} and STL-10~\cite{coates2011analysis} are ten-class image classification datasets.
In addition, we analyze the inherent functionality of NTP on more complex datasets, 
namely ImageNette and ImageWoof~\cite{howard2020fastai}.
ImageNette contains 10 distinct classes from ImageNet~\cite{russakovsky2015imagenet} and Imagewoof is another subset that has high similarity.
For simple datasets, we normalize the input size as $(32,32,3)$, and $(224,224,3)$ for the ImageNet data.

\noindent\textbf{Models:} \label{exp: models}
Following the settings in~\cite{wang2021non, wang2023model}, we implement VGG-11~\cite{simonyan2014very} and ResNet-18~\cite{he2016deep} as backbones for these datasets.
In addition to the source model trained with supervised learning, we also evaluate the self-supervised models, which use SimCLR~\cite{chen2020simple} and MoCo~\cite{he2019moco, chen2020improved} with ResNet-50~\cite{he2016deep} as backbones. 
\ls{??? Unclear to me. You are using the ADMM method for training, right? Did you mean supervised versus self-supervised?}
\ls{Your section 4 is only for supervised learning. Please briefly mention how you can use the framework in section 4 under self-supervised learning -- maybe pseudo-labeling?}
\rd{No, we first have a pretrained model on the source domain, and using ADMM to prune it. The difference between supervised and self-supervised is how the original model is trained.}
\fei{is the last part explained self-supervised? straighten out the last line}
\rd{revised}

\noindent\textbf{Baseline Prior Work and Metrics:} \label{exp: baselines}
We compare our NTP with two baseline prior work,  
NTL\cite{wang2021non} and CUTI~\cite{wang2023model}. 
Their transferability is measured with SLC-AUC.
For self-supervised models, we evaluate the feature space transferability using SOTA metrics: LogME~\cite{you2021logme}  and SFDA~\cite{shao2022not}. 
\ls{If possible, use one sentence to justify why these two?}
\rd{There are only two works by now. Do we need to mention that only two works for NTL?} \ding{If those are the only two prior works, suffice to say so.}

\noindent\textbf{Implementation Details:} \label{exp: implementation details}
During NTP, we utilize $10\%$ samples from the source dataset and target dataset to do model pruning by default, which is further evaluated in Appendix E. 
The ADMM involves an early stop criterion when the ADMM loss converges or reaches the max sparsity (default as $0.99$). \ls{possibly minor??? This stopping criterion does not match Algorithm 1.}
\rd{During the implementation, we will have a hard stop when the model achieve some sparsity, do we need to mention it in algorithm 1?}
During the transferability evaluation, all models are fully fine-tuned for $30$ epochs with Adam optimizer~\cite{kingma2014adam} at a learning rate of $10^{-3}$ with a step scheduler every $10$ epochs by default. 
Ablation study on the learning rate is done in~\cref{tab: different fine-tune}.
Each experiment is repeated for $5$ times with different random seeds and the error bars are visualized.
The batch size is $256$ for model training and fine-tuning.
All experiments are conducted on an NVIDIA TITAN RTX with 24 GB of memory.

\subsection{Evaluation on Supervised Learning}
\label{exp: supervised model}

\begin{table}[t]
\begin{minipage}[t]{0.55\textwidth}
\centering
\caption{Model non-transferability comparison between our proposed \textbf{NTP} with two baselines  {NTL}~\cite{wang2021non} and {CUTI}~\cite{wang2023model} on different Source-Target pairs on VGG-11.}
\label{tab: supervised ntp}
\scriptsize
\begin{tabularx}{\linewidth}{@{}l|X|X|X|X|X@{}}
\toprule
\textbf{\diagbox{Target}{Source} }&  \textbf{\makecell{MT}} &  \textbf{\makecell{UP}} &  \textbf{\makecell{MM}} &  \textbf{\makecell{SN}} &   \textbf{\makecell{SD}} \\ \midrule
\textbf{\makecell{MT}}  \makecell{ {NTL} \\ {CUTI} \\\textbf{{NTP*}}}  &
\makecell{ -} &
\makecell{ {0.42} \\ {0.89} \\\textbf{{-0.47}}} &
\makecell{ {0.41} \\ {1.07} \\\textbf{{-0.07}}} &
\makecell{ {0.35} \\ {0.70} \\\textbf{{-1.02}}} &
\makecell{ {0.47} \\ {0.81} \\\textbf{{-0.64}}} \\
\hline
\textbf{\makecell{UP}}  \makecell{ {NTL} \\ {CUTI} \\\textbf{{NTP*}}} &
\makecell{ {0.23} \\ {2.03} \\\textbf{{-0.47}}} & 
\makecell{ -} &
\makecell{ {0.64} \\ {2.12} \\\textbf{{-1.27}}} &
\makecell{ {0.83} \\ {1.86} \\\textbf{{-1.56}}} & 
\makecell{ {0.16} \\ {1.91} \\\textbf{{-0.78}}} \\
\hline
\textbf{\makecell{MM}}  \makecell{ {NTL} \\ {CUTI} \\\textbf{{NTP*}}}  &
\makecell{ {1.36} \\ {1.49} \\\textbf{{0.16}}} &
\makecell{ {1.32} \\ {1.15} \\\textbf{{-0.13}}} &
\makecell{-} & 
\makecell{ {1.37} \\ {1.24} \\\textbf{{0.01}}} &
\makecell{ {1.19} \\ {1.22} \\\textbf{{-0.06}}} \\
\hline
\textbf{\makecell{SN}}  \makecell{ {NTL} \\ {CUTI} \\\textbf{{NTP*}}}  &
\makecell{ {1.47} \\ {0.72} \\\textbf{{-0.61}}} &
\makecell{ {1.48} \\ {0.68} \\\textbf{{-0.64}}} &
\makecell{ {1.45} \\ {1.03} \\\textbf{{-0.40}}} &
\makecell{ -} &
\makecell{ {1.49} \\ {1.53} \\\textbf{{0.50}}} \\
\hline
\textbf{\makecell{SD}}  \makecell{ {NTL} \\ {CUTI} \\\textbf{{NTP*}}}  & 

\makecell{ {1.52} \\ {0.98} \\\textbf{{-0.34}}} &
\makecell{ {1.56} \\ {0.99} \\\textbf{{-0.25}}} &
\makecell{ {1.64} \\ {1.23} \\\textbf{{-0.46}}} &
\makecell{ {1.52} \\ {1.67} \\\textbf{{0.51}}} &
\makecell{- } \\
\bottomrule
\end{tabularx}
\end{minipage}
\hfill
\begin{minipage}[t]{0.42\textwidth}
\centering
        \caption{Source Accuracy Drop}
    \begin{tabularx}{\linewidth}{@{}X|X|X|X@{}}
    \toprule
    Source  &  NTL  &  CUTI  &  \textbf{NTP*} \\
    \hline
         MT           &{$\mathbf{0.55\%\downarrow}$} & {$0.59\%\downarrow$} & {$0.73\%\downarrow$}  \\
         \hline
         UP           & {$0.11\%\uparrow$} & {$0.16\%\uparrow$} &{$\mathbf{0.31\%\uparrow}$}  \\
         \hline
         MM          &   {$2.57\%\downarrow$} &  {$\mathbf{0.75\%\uparrow}$} & \textbf{{$0.50\%\downarrow$}} \\
         \hline
         SN           &   {$3.50\%\downarrow$} & {$\mathbf{0.17\%\uparrow}$} & \textbf{{$4.10\%\downarrow$}}\\
         \hline
         SD           &   {$1.01\%\downarrow$} & {$\mathbf{0.66\%\uparrow}$} & \textbf{{$0.01\%\uparrow$}}  \\
         \hline
    \end{tabularx}
\vspace{2mm}
    \label{tab: source-domain-accuracy-drop}

        \caption{Fine-tuning Schemes}
    \begin{tabularx}{\linewidth}{@{}l|X|X|X@{}}
    \toprule
    Method/lr  &  \makecell{NTL} & \makecell{CUTI}   &  \makecell{\textbf{NTP*}} \\
    \hline
\textbf{\makecell{FF}}  \makecell{ {$10^{-3}$} \\ {$10^{-4}$} \\\textbf{{$10^{-5}$}}} & \makecell{0.79\\0.73\\0.60}  & \makecell{0.56\\0.95\\0.96}
 & \makecell{\textbf{0.29}\\\textbf{0.32}\\\textbf{0.21}}   \\ 
 \hline
 \textbf{\makecell{LP}}  \makecell{ {$10^{-3}$} \\ {$10^{-4}$} \\\textbf{{$10^{-5}$}}} & \makecell{0.79\\0.69\\0.26}  & \makecell{\textbf{0.18}\\0.73\\\textbf{0.07}} 
 &  
 \makecell{0.28\\\textbf{0.25}\\0.13}\\
         \hline
    \end{tabularx}

    \label{tab: different fine-tune}
\end{minipage}
\end{table}

\ls{Overall, the writing is good. Pay attention to grammar. Also, some writing is a bit wordy. please check}
We compare the proposed NTP with the two baseline approaches. We reproduce NTL\footnote{\url{https://github.com/conditionWang/NTL}} and CUTI\footnote{\url{https://github.com/LyWang12/CUTI-Domain}}, applying them to VGG-11 across various source-target domain pairs. \cref{tab: supervised ntp} shows the non-transferability results, and \cref{tab: source-domain-accuracy-drop} reports the average accuracy on the source domain, showing a trade-off between model non-transferability and source-domain performance for all these methods.

Our findings reveal that NTP surpasses the baseline methods by exhibiting a negative SLC-AUC, indicating that it effectively diminishes the model accuracy on the target domain  to a level even below that achievable by training the model from scratch.
Though both baseline methods report a low target domain performance initially,\fei{what is the context for initiallly? with small dataset, or training epoches?}\rd{these methods has a good performance without fine-tuning} they exhibit less robustness to malicious fine-tuning.\fei{dataset reveal pre-learned knowledge?  isn't the knowledge in the pre-trained model? confusing here.}\rd{Delete the last sentence, I mean that the model accuracy will recover, the same thing as these methods have less robustness to fine-tuning.}
As for the impact on the performance within the source domain, CUTI~\cite{wang2023model} manifests the least impact.
Nonetheless, its effectiveness in restricting transferability is the most limited, particularly when the target domain involves simpler datasets (e.g., UP, MM). 
Our NTP method, while causing a comparable reduction in source domain performance to NTL~\cite{wang2021non}, significantly enhances non-transferability across various domains.
It is important to note that the level of non-transferability also depends on the characteristics of the datasets involved. 
For instance, the SN and SD datasets, both consisting of RGB digits, share considerable similarities. 
This intrinsic similarity between datasets implies that, despite employing a non-transferable learning scheme, completely preventing the leakage of pre-trained information to the target domain remains challenging.
More results about the NTP performance on ResNet-18 can be found in Appendix B.

In~\cref{tab: different fine-tune}, we further evaluate all methods under two transfer learning strategies with three learning rates ($10^{-3}$, $10^{-4}$, $10^{-5}$), and report the target domain accuracy  (\textbf{the lower the better}).
We use the MT dataset as the source and conduct a transfer task to UP on VGG-11 with $10\%$ of the data, which is the common setting of transfer learning.
Full Fine-tuning (FF), indicating all parameters in the model are trainable, which usually has better performance but is computationally intensive.
Linear Probing (LP), is the most commonly used transfer learning method when the early layers of the model are frozen but the final linear layers are fine-tuned during transfer learning.
Our proposed NTP is stable for different learning rates, but NTL and CUTI might achieve high transferability under specific learning rate settings (with high accuracy).
Moreover, although CUTI has lower accuracy during linear probing, it has a bad performance when the entire model can be fine-tuned. 
The proposed NTP shows high non-transferability for both fine-tuning methods.

\subsection{Evaluation on Self-supervised Learning} \label{exp: self-supervised}
\begin{figure}[t]
    \centering
    \includegraphics[width=0.99\linewidth]{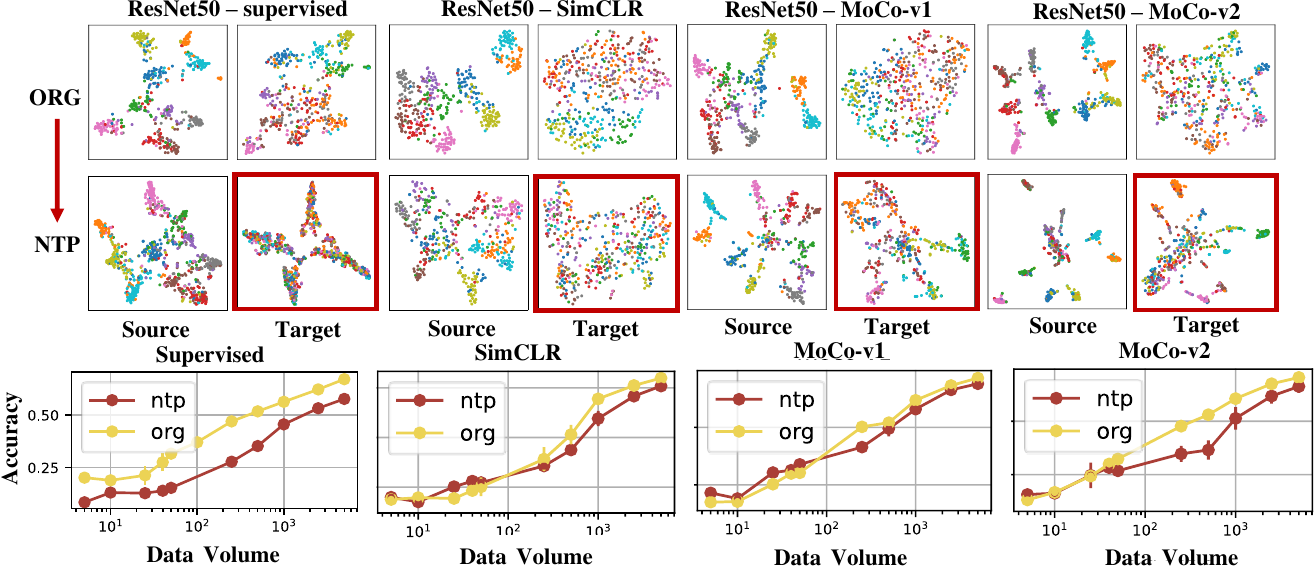}
    \caption{\textbf{Top:} Self-supervised models' feature space with/without NTP using t-SNE~\cite{van2008visualizing}; \textbf{Bottom:} line charts for target accuracy after fine-tuning with various data volumes.}.
    \label{fig: ssl-embeddings}
\end{figure}
Self-supervised Learning (SSL) is declared to offer enhanced generalizability and transfer learning flexibility due to its ability to autonomously extract representations from input data~\cite{ericsson2021well}.
We evaluate NTP across three SSL models—SimCLR~\cite{chen2020simple}, MoCo~\cite{he2019moco}, and MoCo-v2~\cite{chen2020improved}—employing ResNet-50 as the backbone with pretrained checkpoints\footnote{\url{https://github.com/linusericsson/ssl-transfer}} for transferring tasks from CIFAR-10 to STL-10 datasets.
\cref{fig: ssl-embeddings} visualizes the feature spaces via t-SNE for both the source (CIFAR-10) and target (STL-10) datasets, alongside accuracy upon full fine-tuning with varying sample sizes.
As a comparison, we also do the same experiment on a supervised learned model.
NTP can keep the features discrepancy on the source domain, suggesting its ability to maintain source domain performance.
Notably, NTP demonstrates superior performance on the SL models relative to the SSL models in non-transferability.
Particularly, MoCo-v1 and MoCo-v2, still exhibit discernible class clusters for the target domain, indicating some degree of transferability.
In contrast, the features of SL models on the target domain have blurred boundaries and it is more challenging to classify.
\fei{In general, when presenting experimental results, explain the experiments first before you present the results in figures or tables.}

To quantitatively assess the efficacy of NTP in SSL models, we apply two SOTA transferability metrics: the Logarithm of Maximum Evidence (LogME)\cite{you2021logme}; Self-challenging Fisher Discriminative Analysis (SFDA)\cite{shao2022not} in \cref{tab: ssl-metric}.
Higher scores indicate greater transferability.
We demonstrate the effectiveness of NTP--it successfully reduces the transferability scores to STL-10 (target domain) based on these metrics.
Our hypothesis attributes this phenomenon to the intrinsic regularization effect of the SSL training scheme, particularly through contrastive loss, which promotes greater class separation within the feature space.\fei{this paragraph needs more work to explain the results for better understanding. a higher metric means what? what are the source doamin and target domain resepctively?}

\begin{table}[t]
    \centering
    \scriptsize
    \caption{Transferability measurements on self-supervised pretrained model}
    \begin{tabularx}{1\linewidth}{@{}l|X|X|X|X|X|X|X|X|X@{}}
    \toprule
    \multirow{ 2}{*}{\diagbox{Metric}{Model}}   & \multirow{ 2}{*}{\makecell{Domain}}  & \multicolumn{2}{c|}{\makecell{Supervised}}  & \multicolumn{2}{c|}{\makecell{SimCLR~\cite{chen2020simple}}} & \multicolumn{2}{c|}{\makecell{MoCo-v1~\cite{he2019moco}}} & \multicolumn{2}{c}{\makecell{MoCo-v2~\cite{chen2021mocov3}}}\\
    \cline{3-10}
    && \makecell{ORG} & \makecell{NTP} & \makecell{ORG} & \makecell{NTP} & \makecell{ORG} & \makecell{NTP} &  \makecell{ORG} & \makecell{NTP} 
    \\
    \hline
    \multirow{ 2}{*}{LogMe~\cite{you2021logme}}  & \makecell{CF-10} & \makecell{3.52} & \makecell{7.79}& \makecell{7.68} & \makecell{7.45} & \makecell{7.11} & \makecell{7.44} & \makecell{8.56} &  \makecell{8.22}\\  
    & \makecell{STL-10} &  \makecell{3.16} & \makecell{\textbf{0.93}} & \makecell{4.07} &  \makecell{\textbf{0.48}} & \makecell{2.24} & \makecell{\textbf{-2.52}} & \makecell{6.19} & \makecell{\textbf{-1.11}} \\
    \hline
    \multirow{ 2}{*}{SFDA~\cite{shao2022not}} & \makecell{CF-10}  & \makecell{0.96} & \makecell{0.99} & \makecell{0.99} & \makecell{1.0} & \makecell{0.86} & \makecell{0.80} & \makecell{0.99} & \makecell{0.69}\\
    & \makecell{STL-10}  & \makecell{0.97} & \makecell{\textbf{0.81}} & \makecell{0.99} & \makecell{\textbf{0.97}} & \makecell{0.72} & \makecell{\textbf{0.36}} & \makecell{0.85} & \makecell{\textbf{0.41}}\\
    \bottomrule
    \end{tabularx}
    \label{tab: ssl-metric}
\end{table}

\ding{explain: Here, the pretained model is not for intended task yet. It is an encoder whose output features will be used for the task. Specify which task the experiment is aiming at. With/without NTP, the features of different classes are both well-separated into different clusters on the source domain, making them suited for transfer-learning to the intended task. The transferability measurement for what task is shown in Table~\ref{tab: ssl-metric}? In comparison, on target domain, the features from different classes are mixed together, so hard for transfer-learning to the task.}
\rd{Yes, that's correct. The curve plot in Figure 4 is the final result after we use the encoder and fine-tune it on the target domain. It shows that on SSL models, the final non-transferability may not be significant due to the training scheme of using their specific loss}

\subsection{NTP Effectiveness in Complex Tasks} \label{exp: visualization}
\begin{figure}[t]
    \centering
    \includegraphics[width=1\linewidth]{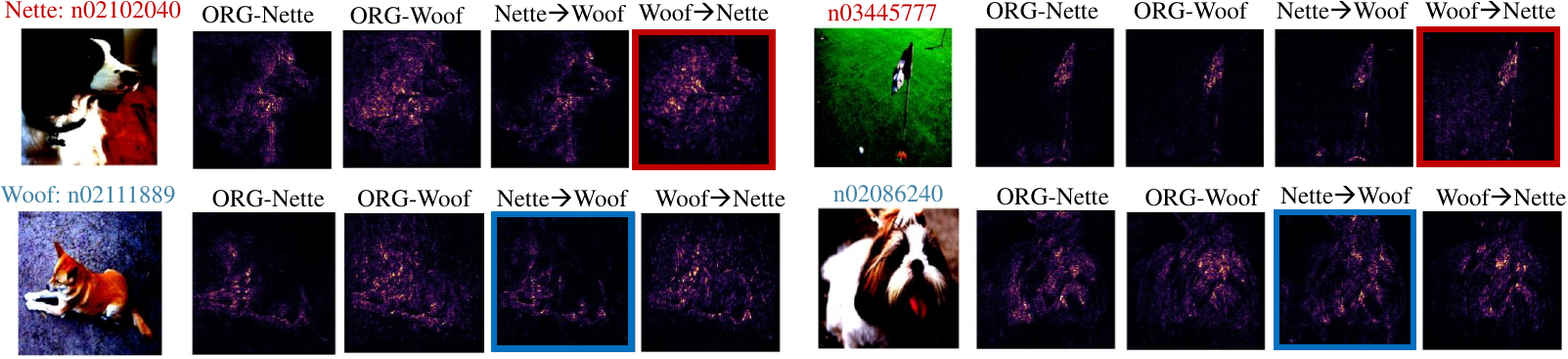}
    \caption{Interpret NTP using Integrated Gradients~\cite{sundararajan2017axiomatic}. Images are evaluated on original/NTP-pruned models between \textbf{\textcolor{darkred}{ImageNette}} and \textbf{\textcolor{darkblue}{ImageWoof}}.
    n02102040, n03445777 are from ImageNette; n02111889, n02086240 are from ImageWoof. } 
    \label{fig: imagenet-visualization}
\end{figure}

\ls{revise the subsection title. Make it clear in the title that you are heading complicated datasets}
\rd{revised}
We use two complex ImageNet subsets, \textcolor{darkred}{ImageNette} and \textcolor{darkblue}{ImageWoof}, to illustrate how NTP curves the model non-transferability.
ImageNette comprises ten classes that are relatively straightforward to classify, whereas ImageWoof includes ten classes of various dog breeds, which are notably more challenging to classify. 
We trained ResNet-18 models on these datasets, achieving $83.3\%$ accuracy on ImageNette and $64.8\%$ on ImageWoof, respectively.
Subsequently, we applied NTP with a desired sparsity of $0.8$ to prune and update the model on ImageNette with restricted transferability to ImageWoof, and vice versa.
When the source domain is ImageNette, the model target accuracy is reduced to $43.7\%$ with a source domain accuracy of $81.7\%$;
For ImageWoof as the source, NTP lowered ImageNette performance to $68.7\%$, with ImageWoof accuracy at $58.6\%$.
To interpret the models' decision-making processes, we utilized integrated gradient visualization~\cite{sundararajan2017axiomatic}, illustrating the critical features influencing the models' predictions, as shown in~\cref{fig: imagenet-visualization}.
Our analysis indicates that the ImageNette-trained model primarily identifies coarse-grained features, such as object outlines.
NTP emphasizes on this tendency, causing the model to favor these broader features and thus diminish its adaptability to ImageWoof, where recognizing finer details, like distinguishing between dog breeds, is essential.
On the other hand, the ImageWoof-trained model concentrates on fine-grained details, such as dogs' eyes and fur.  
Applying NTP heightens this detailed focus, leading to the inclusion of extra noise and a subsequent drop in its ImageNette performance.
\ls{please check grammar and avoid unnecessary lengthy descriptions.}
\rd{revised.}

\subsection{Ablation Studies} \label{exp: hyperparameters}

First, we adjusted NTP's sparsity level, observing (in Appendix C) that higher sparsity correlates with reduced transfer learning performance in the target domain. 
Next, we vary the number of epochs in the weight-updating step in the ADMM algorithm from one to twenty.
We find that even with small weight updating epochs in ADMM, NTP can still achieve low SLC-AUC scores (Appendix D).
We then experimented with varying the training percentage data for ADMM, finding NTP to be more effective with more training samples. 
This suggests that when NTP utilizes a small portion of training dataset, it may overly focus on specific samples, leading to a biased pruning process (Appendix E).
\fei{rephrase this line} \rd{revised}

\section{Conclusions} \label{sec: conclusion}
In this study, we delve into DNN non-transferable learning, an emerging and crucial aspect in safeguarding the IP of DNN models. 
We introduce Non-transferable Pruning, which incorporates an ADMM-based pruning approach and an optimization using Fisher discriminative regularization. 
Furthermore, our research underlines the importance of evaluating model transferability across various fine-tuning settings, especially for potential attackers who may possess full access to the model. 
As the value of DNN intellectual property continues to increase, especially in the case of larger and more complex models, NTP emphasizes the need for proactive protection in model applicability authorization for model vendors.
\noindent\textbf{Limitation.} In implementing ADMM for NTP, we appreciate its advantages in fast convergence and scalability for large-scale models. 
Yet, this approach faces notable challenges. 
First, it shows sensitivity to hyperparameters. 
The integration of non-transferable loss and Fisher space regularization necessitates careful calibration of hyperparameters. 
This is particularly true for setting the maximum loss for $\mathcal{R}_\mathcal{T}$, which is critical to maintaining loss stability. 
Similarly, the regularization weights for $\mathcal{R}_{\Phi_\mathcal{T}}$ demand careful adjustment to strike a balance among different loss terms.
Secondly, there's a delicate balance between model sparsity and preserving non-transferability.
Setting the right sparsity level in NTP is critical to maintain both the model's effectiveness in the source domain and its non-transferability for complex transfer learning tasks.
\ls{Do we need to have both discussion and conclusion sections?}


%
%
\newpage
\section*{Acknowledgements}
This work was supported in part by National Science Foundation under grants SaTC-1929300 and CNS-2212010. 
L. Su is supported in part by an NSF CAREER award CCF-2340482.
 
\bibliographystyle{splncs04}
\bibliography{egbib}
\newpage
\appendix
\section{Proof of Theorem \ref{thm: main}} \label{appendix: fisher-space-regularizer}

Recall that the label set $\mathcal{C}$ of the target domain is binary, $\{0,1\}$, and the that two classes have an equal number of samples.  
We will construct a feature extractor, denoted by $f$, such that the numerator of~\cref{eq: fisher regularizer} is 0, and the denominator of~\cref{eq: fisher regularizer} can be made arbitrarily large. 

Let $|C_1| = |C_0| = m$. Denote the input features of the samples in $C_0$ as 
\begin{align*}
    \{x_1^0, x_2^0, \cdots, x_m^0\},
\end{align*}
and the input features of the samples in $C_1$ as 
\begin{align*}
    \{x_1^1, x_2^1, \cdots, x_m^1\}. 
\end{align*}
%
Without loss of generality, assume that no two inputs $x$ are identical. 
Let $f$ be a function such that
$$
\left\{ \begin{array}{lll}
f(x_1^1) = f(x_1^0) = z_1\\
f(x_2^1) = f(x_2^0)=z_2\\
\dots\\
f(x_m^1) = f(x_m^0)=z_m
\end{array}\right.
$$
It is easy to see that for any of such feature map, we have $\bar{z}_1 = \bar{z}_0$, which immediately leads to 
\begin{align*}
\left \|\overline{z}_0 -\frac{1}{2}(\overline{z}_{0}+ \overline{z}_{1})\right\|_2 + \left \|\overline{z}_1 -\frac{1}{2}(\overline{z}_{0}+ \overline{z}_{1})\right\|_2 =0,     
\end{align*}
minimizing~\cref{eq: fisher regularizer}. 
Furthermore, the denominator 
\[
\sum_{(x_i, y_i)\in C_0} ||z_i -\overline{z}_0||_2 + \sum_{(x_i, y_i)\in C_1} ||z_i -\overline{z}_1||_2 
\]
can be controlled by searching among proper feature mapping $f$ to encourage feature variance. By the universal approximation theorem of neural networks 
, any function $f$ that satisfies mild technical regularity assumptions can be well approximated by neural networks with architectures as simple as sufficiently wide feedforward neural network with one hidden layer. 

 

\section{NTP performance on ResNet-18} \label{appendix: digit-resnet}
\begin{table}[ht]
\centering
    \caption{NTP SLC-AUC scores for ResNet-18 on Digits Datasets}
\begin{tabularx}{0.7\linewidth}{@{}l|X|X|X|X|X@{}}
\toprule
\textbf{\diagbox{Target}{Source} }&  \textbf{\makecell{MT}} &  \textbf{\makecell{UP}} &  \textbf{\makecell{MM}} &  \textbf{\makecell{SN}} &   \textbf{\makecell{SD}} \\ \midrule
\textbf{\makecell{MT}}   &
\makecell{ -} &
\makecell{ -1.02} &
\makecell{-2.07} &
\makecell{ -1.71} &
\makecell{ -2.11} \\
\hline
\textbf{\makecell{UP}}  &
\makecell{ -2.03} & 
\makecell{ -} &
\makecell{ -2.24} &
\makecell{ -2.36} & 
\makecell{ -1.71} \\
\hline
\textbf{\makecell{MM}}   &
\makecell{ -0.91} &
\makecell{ -1.13} &
\makecell{-} & 
\makecell{ -1.49} &
\makecell{-0.64} \\
\hline
\textbf{\makecell{SN}}  &
\makecell{ -0.75} &
\makecell{ -0.72} &
\makecell{ -0.29} &
\makecell{ -} &
\makecell{ 0.33} \\
\hline
\textbf{\makecell{SD}}  & 

\makecell{-0.58} &
\makecell{ -0.69} &
\makecell{-0.18} &
\makecell{-0.07} &
\makecell{- } \\
\bottomrule
\end{tabularx}

    \label{tab: ntp on resnet}
\end{table}
We evaluate the NTP performance on ResNet-18 across different pairs of datasets on ResNet-18.
Note that the baseline methods are only implemented for the VGG structure, but we demonstrate the effectiveness of NTP for ResNet-18.
Similar to the performance for VGG-11, NTP successfully reduces the model transferability to a negative SLC-AUC level except for the source-target pair between the SD (Synthetic Digits) and SVHN dataset.

\section{Various Sparsity in NTP} \label{appendix: pruning-ratio}

\begin{figure}[ht]
    \centering
    \includegraphics[width=0.85\linewidth]{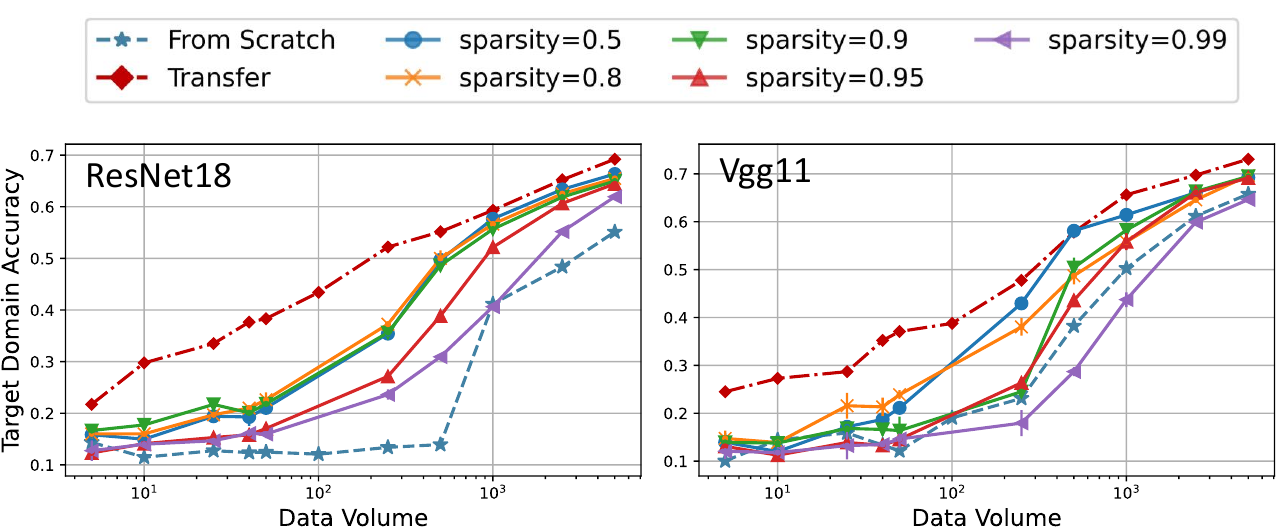}
    \caption{Sample-wise Learning Curves for NTP-pruned model with different sparsity for CIFAR-10 (Source Domain) to STL-10 (Target domain) for ResNet-18 and VGG-11.}
    \label{fig: ntp vs sparsity}
\end{figure}

Model sparsity after the pruning process provides a significant impact on the model's transferability, as illustrated in~\cref{fig: ntp vs sparsity}.
Our experimental results demonstrate sample-wise learning curves for both ResNet18 and VGG11 during the transfer from the CIFAR-10 to STL-10 datasets, employing the NTP method with a hard stop at various sparsity levels. 
The dashed red and blue curves represent the SLC of models utilizing transfer learning and training from scratch, respectively. 
\cref{fig: ntp vs sparsity} shows that NTP is effective for both models and it can always reduce the model's transferability compared with conventional transfer learning.
Specifically, the model's non-transferability of NTP is proportional to its sparsity--a higher sparsity indicates a smaller model capacity, thus reducing its transfer learning performance on the target domain.
Interestingly, we observe that the final accuracy of STL-10 when training the model from scratch is lower than using transfer learning.
The possible reasons include the total number of training samples from STL-10 is relatively small ($5000$), but the pre-learned knowledge from CIFAR10 improves its final performance.

\section{Various Epoch Number in ADMM Weight Updating} \label{appendix: ADMM-epochs}
\begin{figure}[ht]
    \centering
    \begin{minipage}[t]{0.49\textwidth}
    \includegraphics[width=\linewidth]{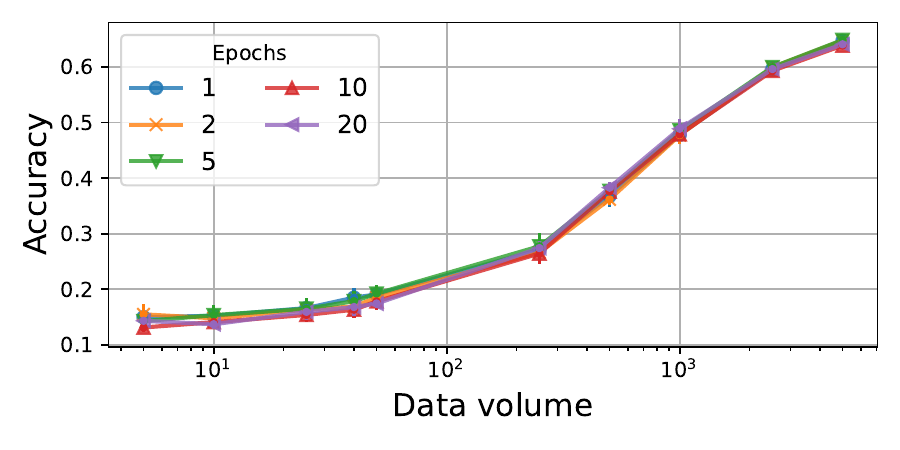}
    \caption{ADMM with Various Number of Epochs During Weight Updating.}
    \label{fig: admm-epoch}
    \hfill
    \end{minipage}
    \begin{minipage}[t]{0.49\textwidth}
    \includegraphics[width=\linewidth]{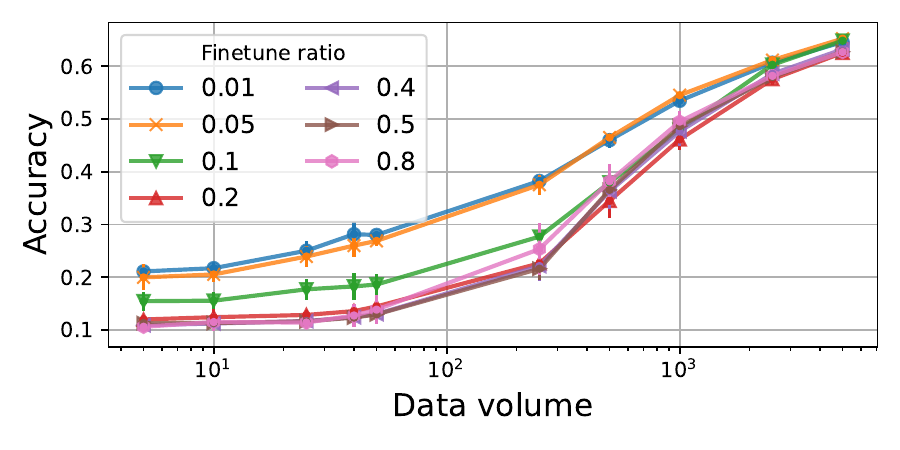}
    \caption{ADMM with Various Training Data Size From CIFAR-10 to STL-10.}
    \label{fig: admm-finetune}
    \hfill
    \end{minipage}

\end{figure}

We evaluate the ADMM performance with various numbers of epochs during weight updating using SGD in the ADMM and we select the source domain as CIFAR-10 and the target domain as STL-10 on ResNet-18, as shown in~\cref{fig: admm-epoch}.
We observe that the selection of the number of epochs has no obvious impact on the final performance.
Update the weights for 10 epochs during every ADMM evaluation shows the best performance, but using an epoch number equal to 1 shows comparable performance with less computational cost.

\section{Various Amount of Data for NTP} \label{appendix: ADMM-data}

We evaluate the number of training data used for NTP in the ADMM pruning procedure on CIFAR-10 to STL-10 on ResNet-18, as shown in~\cref{fig: admm-finetune}.
Using a large number of data to prune the model, we can obtain a higher non-transferability.
Due to the bias introduced by pruning with a small number of samples, the model shows worse performance to control the model transferability.
In our work, we use $10\%$ data for model pruning as a trade-off between computation efficiency and final model non-transferability.
\end{document}